# DeepSeek-R1 Outperforms Gemini 2.0 Pro, OpenAI o1, and o3-mini in Bilingual Complex Ophthalmology Reasoning

Pusheng Xu, MD[1], Yue Wu, MD[1], Kai Jin, MD, PhD[2], Xiaolan Chen, MD[1], Mingguang He, MD, PhD[1, 3, 4], Danli Shi, MD, PhD[1, 3*]

**Affiliations**

1. School of Optometry, The Hong Kong Polytechnic University, Kowloon, Hong Kong.

2. Zhejiang University, Eye Center of Second Affiliated Hospital, School of Medicine, Hangzhou, Zhejiang, China.

3. Research Centre for SHARP Vision (RCSV), The Hong Kong Polytechnic University, Kowloon, Hong Kong.

4. Centre for Eye and Vision Research (CEVR), 17W Hong Kong Science Park, Hong Kong

**Correspondence**

*__Dr Danli Shi,__ MD, PhD., The Hong Kong Polytechnic University, Kowloon, Hong Kong SAR, China.

Email: danli.shi@polyu.edu.hk

**Abstract**

**Purpose:** To evaluate the accuracy and reasoning ability of DeepSeek-R1 and three other recently released large language models (LLMs) in bilingual complex ophthalmology cases.

**Methods:** A total of 130 multiple-choice questions (MCQs) related to diagnosis (n = 39) and management (n = 91) were collected from the Chinese ophthalmology senior professional title examination and categorized into six topics. These MCQs were translated into English using DeepSeek-R1. The responses of DeepSeek-R1, Gemini 2.0 Pro, OpenAI o1 and o3-mini were generated under default configurations between February 15 and February 20, 2025. Accuracy was calculated as the proportion of correctly answered questions, with omissions and extra answers considered incorrect. Reasoning ability was evaluated through analyzing reasoning logic and the causes of reasoning error.

**Results:** DeepSeek-R1 demonstrated the highest overall accuracy, achieving 0.862 in Chinese MCQs and 0.808 in English MCQs. Gemini 2.0 Pro, OpenAI o1, and OpenAI o3-mini attained accuracies of 0.715, 0.685, and 0.692 in Chinese MCQs (all P<0.001 compared with DeepSeek-R1), and 0.746 (P=0.115), 0.723 (P=0.027), and 0.577 (P<0.001) in English MCQs, respectively. DeepSeek-R1 achieved the highest accuracy across five topics in both Chinese and English MCQs. It also excelled in management questions conducted in Chinese (all P<0.05). Reasoning ability analysis showed that the four LLMs shared similar reasoning logic. Ignoring key positive history, ignoring key positive signs, misinterpretation medical data, and too aggressive were the most common causes of reasoning errors.

**Conclusion:** DeepSeek-R1 demonstrated superior performance in bilingual complex ophthalmology reasoning tasks than three other state-of-the-art LLMs. While its clinical applicability remains challenging, it shows promise for supporting diagnosis and clinical decision-making.

**Introduction**

Large language models (LLMs), such as OpenAI GPT series[1] and Google's Gemini series,[2 3] have revolutionized the field of artificial intelligence (AI) by demonstrating impressive capabilities in natural language understanding and reasoning. These models exhibit significant potential in the medical domain, including personalized health consultations, research and clinical decision support, surgical planning assistance, and the facilitation of telemedicine.[4] However, their performance and safety must undergo rigorous evaluation before they can be responsibly integrated into clinical workflows.[5]

In ophthalmology, researchers have tried to use the LLMs to integrate massive amounts of ophthalmic medical literature, guidelines, and patients' data to assist doctors in making more accurate diagnoses and facilitating clinical decision support.[6-12] However, existing LLMs have not yet meet the rigorous standards required for clinical adoption in ophthalmic disease diagnosis.[13] For example, Bahir et al. found that Gemini Advanced only got a 66% accuracy rate in an ophthalmology residency exam.[14] Similarly, Zhang et al. found that GPT-4o exhibited significantly lower accuracy in primary diagnosis compared to human ophthalmologists in twenty-six glaucoma cases.[15] Notably, while LLMs may perform well on recall tasks, they face challenges when handling complex medical cases that require reasoning. For example, GPT-4 achieved mean accuracies of only 48.0% and 63.0% in diagnosing and determining the next step in reasoning tasks derived from JAMA Ophthalmology's Clinical Challenges section.[16]

Recently, the DeepSeek team released its latest cost-effective open-source model, DeepSeek-R1.[17] By incorporating multi-stage training and cold-start data prior to large-scale reinforcement learning (RL), DeepSeek-R1 achieved performance on reasoning tasks comparable to OpenAI-o1-1217. However, its accuracy and clinical applicability in complex ophthalmology reasoning tasks, particularly in a bilingual context, remain uncertain. Bilingual capabilities are essential in clinical ophthalmology, as medical professionals often need to interpret patient records, guidelines, and research findings across multiple languages, especially in multilingual regions or international collaborations. Accurate language comprehension is critical for reducing misinterpretations and ensuring precise diagnoses and treatment recommendations.

This study aims to evaluate the performance of state-of-the-art (SOTA) LLMs, including DeepSeek-R1, Gemini 2.0 Pro, OpenAI o1, and OpenAI o3-mini, in bilingual complex ophthalmology reasoning. By assessing their accuracy and reasoning ability, we seek to determine their potential for real-world clinical applications and identify areas for future improvement.

**Methods**

**Data sources**

To prevent potential data leakage—where test data is used in model training—we didn't use the publicly accessible USMLE questions that had been utilized in previous studies.[18 19] Instead, we collected 130 multiple-choice questions (MCQs) designed for the Chinese ophthalmology senior professional title examination from VIP documents on Baidu Wenku. These MCQs were reviewed for validity and reliability by an ophthalmologist with over six years of clinical experience. The questions assess diagnostic (including differential diagnosis, n=39) and management (n=91) aspects across various ophthalmic subspecialties. We categorized them into six main topics: anterior segment diseases (n=25), external eye/orbital diseases (n=24), glaucoma (n=21), ocular trauma (n=32), refractive disorders/strabismus (n=17), and retinal diseases (n=11). Each question contains 5 to 9 answer choices, with the number of correct answers ranging from 1 to 6. An overview of this study is presented in Figure 1.

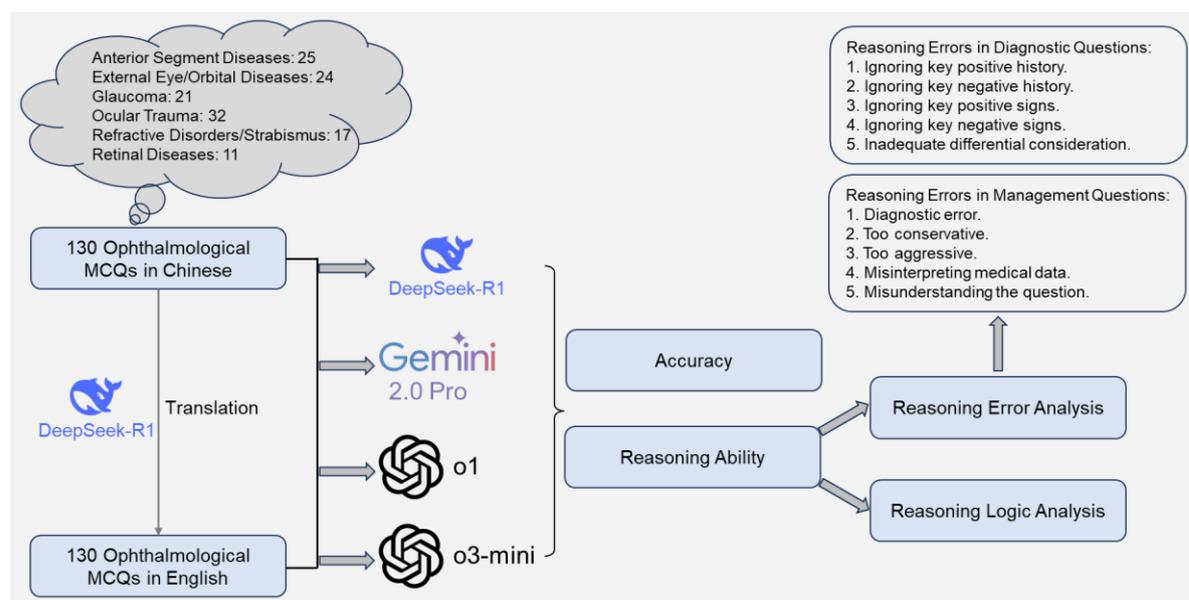

**Figure 1.** Overview of the study.

**Translation of MCQs**

Considering that the same question in different languages may affect the performance of LLMs.[20] We used DeepSeek-R1 to generate the English version of these MCQs. To mitigate the risk of the model unintentionally memorizing the correct answers during translation, only the questions and answer options were input, excluding the reference answers. The prompt used for this translation process is provided in Supplemental Table 1A.

**LLM Access**

We accessed DeepSeek-R1 via the Application Programming Interface (API) provided by Volcengine, a cloud service platform under ByteDance, as the official DeepSeek server has been experiencing attacks, overload pressure, and usage limitations. Gemini 2.0 Pro (Gemini-2.0-pro-exp) and OpenAI o3-mini were accessed through their official APIs. Due to OpenAI's restrictions on API access for o1, we were unable to retrieve responses via the official API and instead obtained them through the official chat user interface (UI).

To ensure linguistic consistency between the responses and the corresponding MCQs, we used prompts aligned with the language of the questions (Supplemental Tables 1B and 1C). All responses were generated under default configurations between February 15 and February 20, 2025.

**Reasoning Ability Analysis**

Reasoning ability was assessed by analyzing reasoning logic and identifying the causes of reasoning errors. The analysis of reasoning logic involved examining and comparing the reasoning processes used to answer questions correctly across all models. The causes of reasoning errors were analyzed separately according to question type. For diagnostic-related questions, errors were classified into five categories: 1. Ignoring key positive history. 2. Ignoring key negative history. 3. Ignoring key positive signs. 4. Ignoring key negative signs. 5. Inadequate differential consideration. For management-related questions, errors were also classified into five categories: 1. Diagnostic error, which means the answer was chosen based on an incorrect diagnosis. 2. Too conservative. For example, ignoring the role of adjunctive therapy or recommending conservative treatment when surgery should be chosen. 3. Too aggressive. For example, recommending a more expensive or higher-risk test or treatment when a cheaper or safer alternative is available. 4. Misinterpreting medical data. For example, misinterpreting disease characteristics and complications, or misunderstanding the indications and contraindications of medications or surgeries. 5. Misunderstanding the question. For example, providing multiple answers when the question asks for the most important or most urgent action.

## Statistical Analysis

The final answers chosen by the LLMs were manually verified based on the text responses of the models. Accuracy was calculated as the ratio of correctly answered questions to the total number of questions. Since some MCQs contained multiple correct answers, both omitted and extra answers were considered incorrect in this study. A 95% confidence interval was calculated using the Clopper-Pearson method. P-values were computed using McNemar's test, with $P < 0.05$ considered statistically significant. Statistical analyses were performed using Stata/MP 17.0 (StataCorp, College Station, TX, USA). Radar charts, grouped bar charts and stacked bar charts were created with Origin 2025 (OriginLab Corporation, Northampton, MA, USA).

## Results

### Overall performance of the four LLMs

As shown in Table 1A, DeepSeek-R1 demonstrated a leading performance in Chinese complex ophthalmology reasoning tasks, achieving an overall accuracy of 0.862 (95% CI: 0.790-0.916; all P<0.001 when compared with three other LLMs). Gemini 2.0 Pro ranked second, with an overall accuracy of 0.715 (95% CI: 0.630-0.791); however, its performance was comparable to OpenAI o1 and o3-mini, as no statistically significant differences were observed.

Although DeepSeek-R1 performed lower accuracy in English reasoning tasks than in Chinese (Table 1B), it still ranked first, achieving an overall accuracy of 0.808 (95% CI: 0.729-0.872; P = 0.115, 0.027 and <0.001 when compared with Gemini 2.0 Pro, OpenAI o1 and o3-mini, respectively). Additionally, both Gemini 2.0 Pro and OpenAI o1 exhibited higher accuracy in English reasoning tasks. In contrast, OpenAI o3-mini demonstrated worse performance in English (P=0.017, Supplemental Table 2), with an accuracy of only 0.577 (95% CI: 0.487-0.633), placing it in fourth position.

**Table 1** | Overall accuracy of DeepSeek-R1 and three other Large Language Models in Chinese (A) and English (B) complex ophthalmology reasoning.

A

| Models | Accuracy | 95% CI | P value |
|---|---|---|---|
| **DeepSeek-R1** | **0.862** | [0.790, 0.916] | - |
| **Gemini 2.0 Pro** | 0.715 | [0.630, 0.791] | <0.001* |
| **OpenAI o1** | 0.685 | [0.597, 0.763] | <0.001* |
| **OpenAI o3-mini** | 0.692 | [0.605, 0.770] | <0.001* |

B

| Models | Accuracy | 95% CI | P value |
|---|---|---|---|
| **DeepSeek-R1** | **0.808** | [0.729, 0.872] | - |
| **Gemini 2.0 Pro** | 0.746 | [0.662, 0.818] | 0.115 |
| **OpenAI o1** | 0.723 | [0.638, 0.798] | 0.027 |
| **OpenAI o3-mini** | 0.577 | [0.487, 0.663] | <0.001* |

* Denotes statistically significant for comparisons against DeepSeek-R1's performance. CI = confidence interval.

**Performance of LLMs in different topics**

In the Chinese MCQs, DeepSeek-R1 achieved the highest accuracy in five topics (**Figure 2A**), including glaucoma (0.952, 95% CI: 0.762-0.999), refractive disorders/strabismus (0.941, 95% CI: 0.713-0.999), external eye/orbital diseases (0.875, 95% CI: 0.676-0.973), ocular trauma (0.843, 95% CI: 0.672-0.947), and anterior segment diseases (0.840, 95% CI: 0.639-0.955). However, only statistically significant when compared with Gemini 2.0 Pro in the topic of glaucoma (Supplemental Table 3A). Gemini 2.0 Pro achieved the highest accuracy in retinal disease topic with an accuracy of 0.727 (95% CI: 0.390-0.940).

In the English MCQs, DeepSeek-R1 also ranked first in five topics (**Figure 2B**): refractive disorders/strabismus (0.941, 95% CI: 0.713-0.999), glaucoma (0.905, 95% CI: 0.696-0.988), external eye/orbital diseases (0.875, 95% CI: 0.676-0.973), ocular trauma (0.781, 95% CI: 0.600-0.907), and anterior segment diseases (0.720, 95% CI: 0.506-0.879). However, statistical significance was only observed when compared with OpenAI o3-mini in the topics of glaucoma and ocular trauma (Supplemental Table 3B). Gemini 2.0 Pro shared the highest accuracy in three topics, while OpenAI o1 shared the highest accuracy in two topics. All models exhibited poor performance in retinal diseases, with OpenAI o3-mini achieving the lowest accuracy of 0.182 (95% CI: 0.023-0.518).

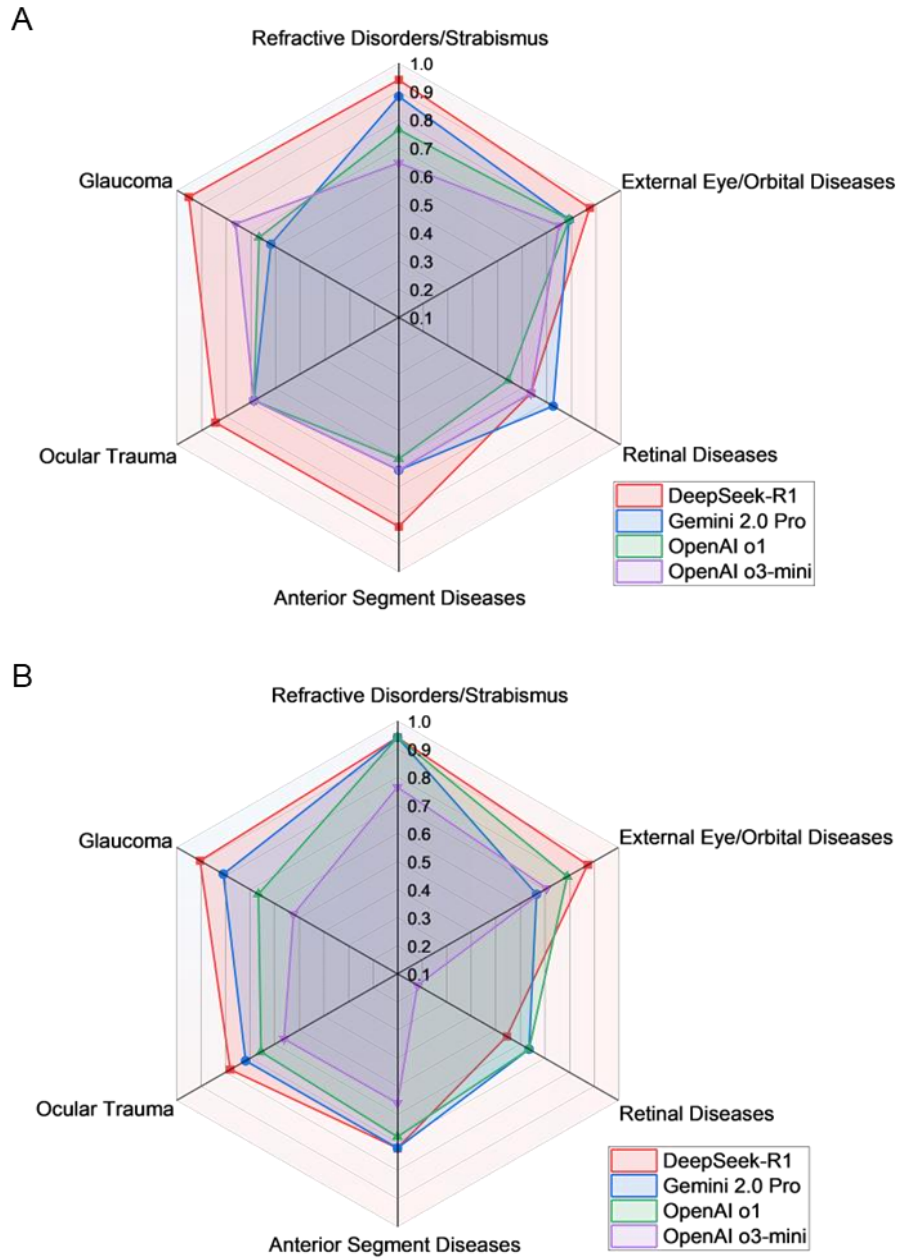

**Figure 2.** Radar charts depicting the performance of the four LLMs across different topics. (A) Chinese MCQs, (B) English MCQs.

**Performance of LLMs in diagnostic and management questions**

As shown in **Figure 3A**, DeepSeek-R1 exhibits superior performance compared to OpenAI o3-

mini and achieves comparable results to Gemini 2.0 Pro and OpenAI o1 in bilingual diagnostic questions. Besides, in management questions conducted in Chinese, DeepSeek-R1 outperforms the three other LLMs, with all comparisons reaching statistical significance (**Figure 3B**).

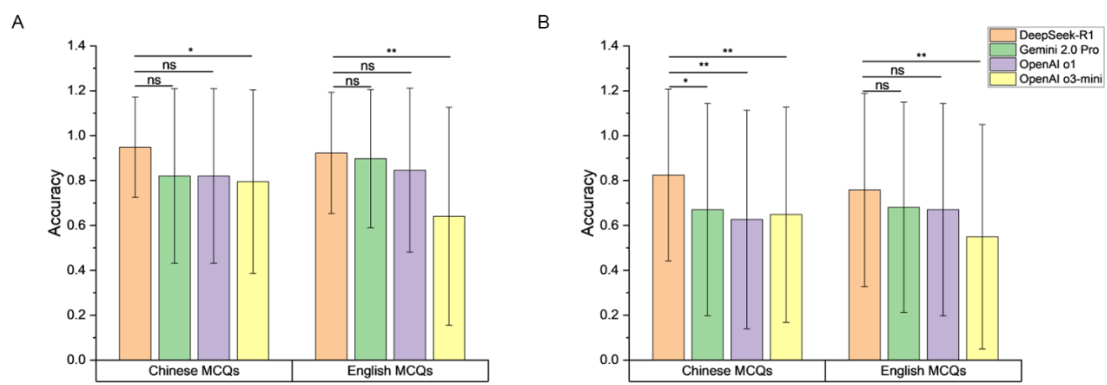

**Figure 3.** Grouped bar charts showing the comparative performance of four LLMs in reasoning (A) diagnostic and (B) management questions.

**Reasoning logic analysis**

All four LLMs correctly answered the same 63 questions in both Chinese and English MCQs. An example is provided in Supplemental Table 4. All models exhibited similar reasoning logic. First, they identified a history of herpes zoster as the most critical clue, recognizing it as a known causative factor of acute retinal necrosis (ARN). Second, they analyzed key positive clinical signs supporting the diagnosis of ARN. Finally, they systematically ruled out incorrect options by eliminating alternative diagnoses.

There were four questions that only DeepSeek-R1 answered correctly in both Chinese and English MCQs. An example is presented in Supplemental Table 5. Although DeepSeek-R1, Gemini 2.0 Pro, and OpenAI o1 recognized that the key to this question was differentiating between an inflammatory pseudotumor of the lacrimal gland and acute dacryoadenitis, their diagnostic approaches differed. DeepSeek-R1 selected blood routine test and ocular ultrasound, whereas Gemini 2.0 Pro and OpenAI o1 opted for magnetic resonance imaging (MRI). In contrast, OpenAI o3-mini deviated by initially focusing on the exclusion of Graves' ophthalmopathy.

**Reasoning error analysis**

Diagnostic errors resulting from ignoring key positive history and positive signs ranked the top-

2 across all LLMs in both Chinese and English diagnostic questions (**Figures 4A and 4B**). In bilingual management questions, misinterpretation of medical data and too aggressive were the two most common errors across all LLMs (**Figures 4C and 4D**).

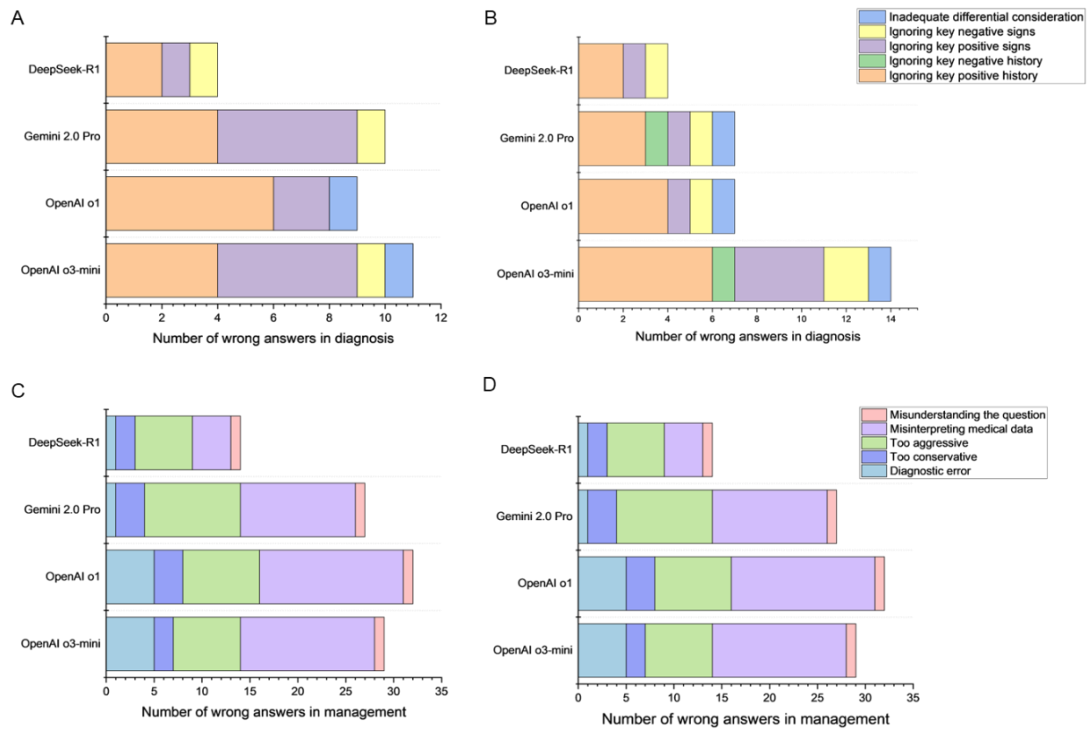

**Figure 4.** Stacked bar chart illustrating the underlying reasons for reasoning errors in LLMs across different question types: (A) diagnostic questions in Chinese MCQs, (B) diagnostic questions in English MCQs, (C) management questions in Chinese MCQs, and (D) management questions in English MCQs.

**Discussion**

Compared to the three other SOTA LLMs, DeepSeek-R1 achieved the best performance in Chinese complex ophthalmology reasoning tasks and performed comparably to Gemini 2.0 Pro in English. DeepSeek-R1 also had the highest number of topics with the highest accuracy and excelled in management questions conducted in Chinese. Reasoning ability analysis showed that the four LLMs shared similar reasoning logic. Ignoring key positive history, ignoring key positive signs, misinterpretation of medical data, and being too aggressive were the most common causes of reasoning errors across all LLMs.

In this study, DeepSeek-R1 demonstrated excellent performance in bilingual complex ophthalmology reasoning tasks, which may be attributed to its innovative training methodology. The DeepSeek team incorporated thousands of high-quality Chain-of-Thought (CoT) data as cold-start data. They observed that fine-tuning the model with cold-start data at the initial phase of reinforcement learning (RL) significantly improved the readability of its output.[17] Additional strategies, including reasoning-oriented RL, rejection sampling, and supervised fine-tuning, were also implemented in DeepSeek-R1. It is likely that this innovative training approach has enabled DeepSeek-R1 to excel in reasoning tasks such as those in this dataset, which require complex and extended logical chains. Several preprint studies have compared the performance of DeepSeek-R1 and other LLMs in medical contexts. For instance, Mondillo et al. reported that OpenAI o1 outperformed DeepSeek-R1 in pediatric MCQs, achieving a diagnostic accuracy of 92.8% compared to 87.0%.[21] Zhou et al. found that DeepSeek-R1 generated more readable responses than ChatGPT-4o in patient education materials for spinal surgeries.[22] Mikhail et al. observed that DeepSeek-R1 and OpenAI o1 demonstrated comparable performance on an English ophthalmology MCQ dataset collected from StatPearls.[23] However, the questions in their dataset consist of one correct answer and three distractor options, making them less challenging than those in our dataset. Furthermore, as their dataset is publicly available, it may have been utilized in the pre-training or fine-tuning of one or more models. To the best of our knowledge, this is the first study to evaluate bilingual complex ophthalmology reasoning performance across DeepSeek-R1 and three other SOTA LLMs.

DeepSeek-R1 performed slightly better in Chinese than in English on our MCQs, which may be attributed to its higher proportion of Chinese pretraining data. Although DeepSeek-R1 has not disclosed the exact proportion of Chinese and English data, its earlier version, DeepSeek-V2, contained 1.12 times more Chinese tokens than English tokens.[24] Gemini 2.0 Pro and OpenAI o1 exhibited superior performance in English, a finding consistent with previous research.[25 26] However, the reason for OpenAI o3-mini's poorer performance in English MCQs remains unclear. In this study, the accuracy of OpenAI o1 was slightly lower than reported in previous studies,[23 27] which may be attributed to the higher difficulty level of the questions used. For instance, in the topic of retinal diseases, where all models demonstrated suboptimal performance, the 11 MCQs included varying numbers of answer choices: 3 questions had five options, 3 had six options, and 5 had eight options. Additionally, 4 questions had more than two correct answers, further increasing the complexity of the task.

All models demonstrated strong reasoning abilities and exhibited similar analytical logic in ophthalmology case analysis. For instance, in diagnostic questions, they first summarize the medical history to identify key clues. Next, they highlight critical positive clinical signs and integrate these with the medical history to formulate a preliminary diagnosis along with a rationale. They then conduct a differential diagnosis for each option, systematically analyzing both supporting and non-supporting factors. Finally, they determine the most appropriate final answer (Supplemental Figure 1). This logical reasoning sequence in case analysis closely aligns with the diagnostic approach used by human physicians. Even in cases where errors in reasoning occurred, they were not due to fundamental flaws in logical structure but rather a failure to identify the most critical clue or select the most appropriate method. For example, as shown in Supplemental Table 5, DeepSeek-R1, Gemini 2.0 Pro, and OpenAI o1 correctly recognized that the key to the question was differentiating between an inflammatory pseudotumor of the lacrimal gland and acute dacryoadenitis. Although the reference answer aligns with DeepSeek-R1's choice of blood routine tests and ocular ultrasound, it is undeniable that the MRI selected by Gemini 2.0 Pro and OpenAI o1 can also effectively differentiate between the two conditions, but its higher cost prevents it from being the first-line choice.[28]

Ignoring key positive history and positive signs was found to be the top two sources of reasoning error. Interestingly, this result aligns with the causes of diagnostic errors observed in human clinicians.[29] Misinterpretation of medical data and recommendation of too aggressive measures may result from wrong information in the training data,[30] variations in reference standards, or the dynamic evolution of medical guidelines. For example, when answering the diagnostic criteria for dry eye using Schirmer's test, the four LLMs failed to reach a consistent conclusion on whether the threshold should be a filter paper wetting length of less than 10mm or less than 5mm within 5 minutes, which may be related to changes in the diagnostic standards for dry eye.[31 32]

This study has several limitations. First, similar to other studies comparing the performance of different LLMs in ophthalmology,[16 23 33] the MCQs used in this study were published before the models' knowledge cutoff date, making it impossible to ensure that these questions were not included in the models' training data. However, the questions were sourced from VIP documents, and Baidu Wenku has implemented various anti-crawling measures for such documents, including asynchronous loading and data encryption, which reduce the likelihood of these documents being included in the training data. Second, due to the lack of reference answers for the reasoning process, we did not quantify the models' reasoning ability. In studies where human answers serve as reference reasoning processes, emerging metrics such as consistency (invariance to logically equivalent inputs), generalization (performance on out-of-distribution data), and explainability (clarity of reasoning steps) can be measured.[34] Although deductive reasoning, inductive reasoning, abductive reasoning, and analogical reasoning are all crucial to the reasoning capabilities of LLMs, abductive reasoning is more commonly used in the medical domain.[35] Therefore, future medical research could prioritize the evaluation of this specific ability.

In summary, compared to the three other LLMs, DeepSeek-R1 exhibited the best performance in bilingual complex ophthalmology reasoning tasks. Although its direct application in clinical practice remains challenging, it holds significant potential for assisting in diagnosis and supporting clinical decision-making.

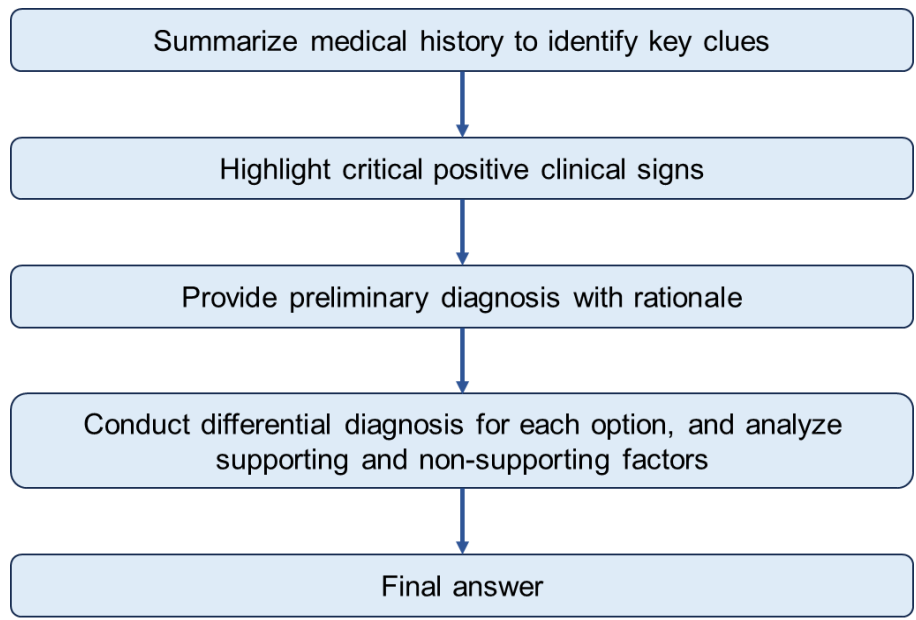

**Supplemental Figure 1.** Logical reasoning framework for diagnostic questions, as shared by all four LLMs.

Supplemental Table 1 | Prompts used for translating Chinese ophthalmology multiple-choice question into English (A), generating responses from Large Language Models in Chinese (B) and English (C).

| List | Prompt |
| --- | --- |
| A | Please translate the following Chinese ophthalmology multiple-choice question into English. The requirements are as follows: <br> 1. The translation must maintain professionalism, ensuring that medical terminology and specialized terms are accurate and precise. <br> 2. Keep the format and structure of the original options unchanged. <br> 3. If the question includes specific clinical or ophthalmological background descriptions, ensure that the translated statements are equally rigorous and fluent in English. <br> 4. For any phrases that may cause ambiguity, please express them clearly and professionally in English. Please translate the following question: {question}. |
| B | 你是一个专业的眼科医生，请分析以下病例，并从给定的选项中选出正确答案：{question}。 |
| C | You are an expert in ophthalmology. Please analyze the following case and select the correct answer(s) from the given options: {question}. |

Supplemental Table 2 | Comparison of the overall performance of 4 LLMs in Chinese and English MCQs.

| Models | Accuracy | | P value |
|---|---|---|---|
| | Chinese MCQs | English MCQs | |
| DeepSeek-R1 | 0.862 | 0.808 | 0.092 |
| Gemini 2.0 Pro | 0.715 | 0.746 | 0.585 |
| OpenAI o1 | 0.685 | 0.723 | 0.473 |
| OpenAI o3-mini | 0.692 | 0.577 | 0.017* |

* Denotes statistically significant.

Supplemental Table 3 | Subgroup analysis of DeepSeek-R1 and three other LLMs' performance in different topics. (A) Chinese MCQs, (B) English MCQs.

A

| Topic | Models | | | |
|---|---|---|---|---|
| | DeepSeek-R1 | Gemini 2.0 Pro | OpenAI o1 | OpenAI o3-mini |
| Anterior Segment Diseases | 0.840 (0.639, 0.955) | 0.640 (0.425, 0.820) | 0.600 (0.387, 0.789) | 0.640 (0.425, 0.820) |
| External Eye/Orbital Diseases | 0.875 (0.676, 0.973) | 0.792 (0.578, 0.929) | 0.792 (0.578, 0.929) | 0.750 (0.533, 0.902) |
| Glaucoma | 0.952 (0.762, 0.999) | 0.619 (0.384, 0.819)* | 0.667 (0.430, 0.854) | 0.762 (0.500, 0.839) |
| Ocular Trauma | 0.843 (0.672, 0.947) | 0.688 (0.500, 0.839) | 0.688 (0.500, 0.839) | 0.688 (0.500, 0.839) |
| Refractive Disorders/Strabismus | 0.941 (0.713, 0.999) | 0.882 (0.636, 0.985) | 0.764 (0.501, 0.932) | 0.647 (0.383, 0.858) |
| Retinal Diseases | 0.636 (0.308, 0.891) | 0.727 (0.390, 0.940) | 0.545 (0.234, 0.833) | 0.636 (0.308, 0.891) |

B

| Topic | Models | | | |
|---|---|---|---|---|
| | DeepSeek-R1 | Gemini 2.0 Pro | OpenAI o1 | OpenAI o3-mini |
| Anterior Segment Diseases | 0.720 (0.506, 0.879) | 0.720 (0.506, 0.879) | 0.680 (0.465, 0.851) | 0.560 (0.349, 0.756) |
| External Eye/Orbital Diseases | 0.875 (0.676, 0.973) | 0.667 (0.447, 0.844) | 0.792 (0.578, 0.929) | 0.708 (0.489, 0.874) |
| Glaucoma | 0.905 (0.696, 0.988) | 0.810 (0.581, 0.946) | 0.667 (0.430, 0.854) | 0.524* (0.298, 0.743) |
| Ocular Trauma | 0.781 (0.600, 0.907) | 0.719 (0.533, 0.863) | 0.656 (0.468, 0.814) | 0.563* (0.377, 0.736) |
| Refractive Disorders/Strabismus | 0.941 (0.713, 0.999) | 0.941 (0.713, 0.999) | 0.941 (0.713, 0.999) | 0.765 (0.501, 0.932) |
| Retinal Diseases | 0.545 (0.234, 0.833) | 0.636 (0.308, 0.891) | 0.636 (0.308, 0.891) | 0.182 (0.023, 0.518) |

* Denotes statistically significant for comparisons against DeepSeek-R1's performance. Data were shown with accuracy (95% confidence interval).

Supplemental Table 4 | Selected examples where all models responded correctly to the same question.

| Question | Female, 32 years old. Complains of redness, pain, and blurred vision in the right eye for 1 week. History of herpes zoster (lumbar region) 1 month prior.<br>Ophthalmic Examination:<br>- Visual acuity: Right eye 0.4 (uncorrected); Left eye 1.2.<br>- Right eye findings:<br>  - Ciliary congestion.<br>  - Mutton-fat keratic precipitates (KP) on corneal endothelium.<br>  - Tydall's phenomenon (positive).<br>  - Pupils isochoric and round; lens clear.<br>  - Moderate inflammatory cells present in the vitreous.<br>- Fundus examination:<br>  - Mild optic disc hyperemia with distinct margins.<br>  - Mild retinal edema.<br>  - Attenuated retinal arteries; peripheral and mid-peripheral arteries exhibit white thread-like appearance with sheathing.<br>  - Patchy yellow-white exudates and scattered dot/patchy hemorrhages in the peripheral retina.<br><br>[Multiple-choice Questions]<br>The most likely diagnosis for this patient is: |
|---|---|
| **Options** | A. Behcet's syndrome<br>B. Acute retinal necrosis **(CORRECT)**<br>C. Toxoplasmic chorioretinitis<br>D. Panuveitis<br>E. Pars planitis<br>F. Collagen vascular disease |
| **Response (Deepseek R1)** | The most likely diagnosis for this patient is **B. Acute retinal necrosis (ARN)**.<br><br>### Key Analysis:<br>1. **History of Herpes Zoster (VZV):**<br>   The patient's recent herpes zoster infection (VZV) strongly suggests a viral etiology. ARN is a rapidly progressive necrotizing retinitis most commonly caused by herpesviruses, including VZV, HSV-1, and HSV-2.<br><br>2. **Ophthalmic Findings Supporting ARN:**<br>   - **Anterior Uveitis:** Ciliary congestion, mutton-fat keratic precipitates (KPs), and Tydall's phenomenon indicate anterior chamber inflammation.<br>   - **Vitritis:** Moderate inflammatory cells in the vitreous suggest posterior segment involvement.<br>   - **Retinal Pathology:**<br>     - Peripheral retinal necrosis (evidenced by patchy yellow-white exudates and hemorrhages).<br>     - Retinal vasculitis (attenuated arteries with white thread-like sheathing).<br>     - Optic disc hyperemia and retinal edema.<br>   These findings align with the classic triad of ARN: **necrotizing retinitis, occlusive vasculitis, and vitritis**. |
|   | 3. **Distinguishing from Other Options:**<br>   - **A. Behcet's Syndrome:** Typically presents with oral/genital ulcers, hypopyon, and retinal vasculitis, none of which are mentioned here.<br>   - **C. Toxoplasmic Chorioretinitis:** Usually manifests as focal retinitis adjacent to old scars, not diffuse peripheral necrosis.<br>   - **D. Panuveitis:** While the patient does have panuveitis (inflammation in anterior, intermediate, and posterior segments), this is a descriptive term rather than a specific diagnosis. ARN is the underlying cause of the panuveitis in this case.<br>   - **E. Pars Planitis:** Characterized by intermediate uveitis with snowbanking, absent here.<br>   - **F. Collagen Vascular Disease:** Retinal vasculitis can occur in lupus or other autoimmune diseases, but the history of VZV makes ARN more likely.<br><br>### Conclusion:<br>The clinical presentation, fundus findings, and recent VZV infection strongly point to **B. Acute Retinal Necrosis** as the specific diagnosis. While panuveitis (D) is present, ARN is the definitive etiology driving the inflammation.<br><br>**Answer:** B. Acute retinal necrosis |
| **Response (Gemini 2.0 PRO)** | Let's break down this case step-by-step, considering the key findings and differential diagnoses:<br><br>**Key Findings:**<br><br>*   **Recent Herpes Zoster:** This is the most crucial piece of information. The history of herpes zoster (shingles) one month prior, even in the lumbar region, strongly suggests a varicella-zoster virus (VZV) etiology.<br>*   **Unilateral Involvement:** The symptoms and signs are confined to the right eye.<br>*   **Anterior Segment Inflammation:**<br>    *   Ciliary congestion: Indicates anterior segment inflammation (uveitis).<br>    *   Mutton-fat KPs: These are large, greasy-appearing precipitates on the corneal endothelium, characteristic of *granulomatous* uveitis. This is a very important clue.<br>    *   Tyndall's phenomenon (positive): Indicates the presence of inflammatory cells in the anterior chamber, confirming anterior uveitis.<br>*   **Posterior Segment Inflammation:**<br>    *   Inflammatory cells in the vitreous: Indicates vitritis.<br>    *   Mild optic disc hyperemia: Suggests optic nerve involvement (but the margins are distinct, ruling out severe optic neuritis).<br>    *   Mild retinal edema.<br>    *   Attenuated retinal arteries with sheathing: This is a very significant finding, pointing towards retinal vasculitis.<br>    *   Peripheral yellow-white exudates and hemorrhages: These are indicative of retinal necrosis and inflammation. The *peripheral* location is very important. |

*Reduced VA: Vision is significantly reduced in the effected eye.

**Differential Diagnosis Analysis:**

*   **A. Behcet's Syndrome:** Behcet's can cause panuveitis, but it's usually bilateral, recurrent, and often associated with oral and genital ulcers. The mutton-fat KPs and strong history of recent zoster make this less likely. Behcet's *can* cause retinal vasculitis, but the other features don't fit as well.

*   **B. Acute Retinal Necrosis (ARN):** This is the *most likely* diagnosis. ARN is a severe, necrotizing retinitis most commonly caused by VZV (or HSV). The key features that point to ARN are:
    *   **History of Herpes Zoster:** ARN is often (but not always) seen in patients with a prior history of zoster or chickenpox.
    *   **Granulomatous Anterior Uveitis:** The mutton-fat KPs suggest a granulomatous inflammatory process, which is consistent with viral uveitis.
    *   **Retinal Vasculitis:** The attenuated, sheathed arteries are a hallmark of ARN.
    *   **Peripheral Retinal Necrosis:** The peripheral yellow-white exudates and hemorrhages represent areas of retinal necrosis. The *peripheral* location is classic for ARN, which typically starts in the periphery and progresses centrally.
    *   **Vitritis:** The presence of inflammatory cells in the vitreous is common in ARN.

*   **C. Toxoplasmic Chorioretinitis:** Toxoplasmosis typically presents with a focal, well-defined area of retinochoroiditis ("headlight in the fog" appearance due to vitritis overlying the lesion). It *can* cause KPs, but mutton-fat KPs are less typical. The history of zoster and the pattern of peripheral necrosis with vasculitis don't fit toxoplasmosis.

*   **D. Panuveitis:** While the patient *does* have panuveitis (inflammation of the anterior uvea, vitreous, and retina), this is a descriptive term, not a specific diagnosis. We need to identify the *cause* of the panuveitis.

*   **E. Pars Planitis:** Pars planitis (intermediate uveitis) is characterized by inflammation primarily in the vitreous and peripheral retina, often with "snowbanking" over the pars plana. While vitritis is present, the presence of mutton-fat KPs, retinal vasculitis, and the history of zoster point away from pars planitis. Pars planitis is also usually bilateral.

*   **F. Collagen Vascular Disease:** While collagen vascular diseases (like lupus or rheumatoid arthritis) *can* cause uveitis and retinal vasculitis, they are much less likely given the acute onset, unilateral presentation, and strong history of recent herpes zoster. The mutton-fat KPs also argue against this.

**Conclusion:**

Based on the clinical presentation, the history of recent herpes zoster,

| | |
|---|---|
| | the presence of granulomatous anterior uveitis, retinal vasculitis, and peripheral retinal necrosis, the most likely diagnosis is **Acute Retinal Necrosis (ARN)** due to reactivation of varicella-zoster virus.

**Final Answer:** The correct answer is **B. Acute retinal necrosis**. |
| **Response (OPENAI o1)** | The most likely diagnosis for this patient is **B. Acute Retinal Necrosis (ARN)**.

**Explanation:**

**Key Clinical Findings:**

1. **History of Herpes Zoster:** The patient had herpes zoster (shingles) in the lumbar region one month prior. Varicella-zoster virus (VZV), which causes shingles, is a known causative agent of ARN.

2. **Ophthalmic Signs:**
   - **Anterior Segment Inflammation:** Ciliary congestion, mutton-fat keratic precipitates, positive Tyndall's phenomenon indicate anterior uveitis.
   - **Posterior Segment Involvement:**
     - **Vitreous Inflammation:** Moderate inflammatory cells in the vitreous.
     - **Retinal Findings:**
       - **Peripheral Retinal Necrosis:** Patchy yellow-white exudates in the peripheral retina.
       - **Retinal Vasculitis:** Attenuated retinal arteries with white thread-like sheathing.
       - **Retinal Hemorrhages:** Scattered dot/patchy hemorrhages in the peripheral retina.
       - **Optic Disc Hyperemia and Mild Retinal Edema:** Indicate posterior segment inflammation.

3. **Unilateral Presentation:** ARN typically presents unilaterally, especially in immunocompetent individuals.

**Why Acute Retinal Necrosis is the Best Fit:**

- **Causative Agents:** ARN is commonly caused by herpes viruses, particularly VZV and HSV.
- **Clinical Features of ARN:**
  - Rapidly progressive, necrotizing retinitis starting in the peripheral retina.
  - Prominent vitritis (inflammatory cells in the vitreous).
  - Occlusive retinal vasculitis (arterial sheathing).
  - Optic nerve involvement and uveitis.
- **Temporal Relationship:** The onset of ocular symptoms one month after a herpes zoster outbreak aligns with the typical progression of VZV reactivation leading to ARN.

**Other Options Considered:**

- **A. Behçet's Syndrome:** Usually presents with recurrent oral and |

| | genital ulcers and bilateral ocular involvement. Retinal vasculitis can occur but is less likely given this patient's history and unilateral findings. |
| | |
| | - **C. Toxoplasmic Chorioretinitis:** Typically presents with focal necrotizing retinochoroiditis ("headlights in fog") and is often associated with congenital infection or immunosuppression. |
| | |
| | - **D. Panuveitis:** While the patient has both anterior and posterior segment inflammation, panuveitis is a broad term. ARN is a specific diagnosis that explains all findings. |
| | |
| | - **E. Pars Planitis:** A type of intermediate uveitis affecting the vitreous and peripheral retina, but usually bilateral and without the severe retinal necrosis seen in ARN. |
| | |
| | - **F. Collagen Vascular Disease:** Can cause retinal vasculitis but is less likely without systemic symptoms and given the recent VZV infection. |
| | |
| | **Conclusion:** |
| | |
| | Given the patient's recent VZV infection, unilateral ocular findings consistent with retinal necrosis, vasculitis, and both anterior and posterior segment inflammation, **Acute Retinal Necrosis** is the most probable diagnosis. |
| | |
| | **Answer:** **B. Acute retinal necrosis** |
| **Response (OPENAI o3-mini)** | Based on the given case details, the most likely diagnosis is **B. Acute retinal necrosis**. |
| | |
| | Here's the reasoning: |
| | - The patient has a history of herpes zoster, which is a known risk factor for developing **acute retinal necrosis (ARN)**. ARN typically occurs in immunocompetent patients with a recent history of viral infection, especially herpes zoster or herpes simplex virus. |
| | - The key findings, such as **mutton-fat keratic precipitates**, **positive Tydall's phenomenon**, and **vitreous inflammation**, point to an intraocular inflammation (uveitis). |
| | - The **retinal changes**, such as **white thread-like sheathing of the arteries**, **patchy yellow-white exudates**, and **dot/patchy hemorrhages** in the peripheral retina, are highly suggestive of **retinal necrosis** associated with viral infections like herpes zoster. |
| | - ARN can cause significant vision loss if not treated promptly. |
| | |
| | Other options: |
| | - **A. Behcet's syndrome**: Behcet's disease is a multi-system inflammatory disorder, but the specific retinal findings are not typical of this condition. |
| | - **C. Toxoplasmic chorioretinitis**: This usually presents with more localized retinal lesions and is associated with immunocompromised states. |
| | - **D. Panuveitis**: While panuveitis could cause similar symptoms, the retinal findings and the herpes zoster history make ARN more |

| | likely.
- **E. Pars planitis**: This typically affects younger individuals and is associated with non-infectious inflammation of the peripheral retina, which does not fit the case.
- **F. Collagen vascular disease**: This could cause uveitis but would likely involve other systemic symptoms that are not present here.

Thus, **acute retinal necrosis (ARN)** is the most likely diagnosis. |

Supplemental Table 5 | Selected examples where only Deepseek R1 responded correctly to the same question.

| Question | Female, 45 years old, presents with bilateral eyelid swelling and orbital pain for half a month. Examination findings: Visual acuity OD 1.0, OS 1.0. Both upper eyelids exhibit an S-shaped curvature with erythematous and swollen skin. Tenderness localized to the lateral upper eyelids. Swollen palpebral lacrimal glands observed in the superotemporal fornix. Lateral bulbar conjunctival congestion and edema (+). Corneas transparent. Fundus examination unremarkable. Vital signs: temperature 36.5°C, blood pressure 130/80 mmHg. History of recent cold. No enlargement of superficial lymph nodes.<br><br>[Multiple-choice Question] Further investigations required for definitive diagnosis include: |
|---|---|
| **Options** | A. Blood routine test + ocular ultrasound **(CORRECT)**<br>B. Orbital CT<br>C. MRI<br>D. Thyroid function tests<br>E. Blood routine test |
| **Response (Deepseek R1)** | The patient's presentation of bilateral S-shaped eyelid curvature, erythema, tenderness over the lateral upper eyelids, and swollen palpebral lacrimal glands in the superotemporal fornix strongly suggests acute dacryoadenitis. The bilateral involvement and recent viral illness point to a potential viral etiology, though bacterial or systemic inflammatory causes (e.g., sarcoidosis, IgG4-related disease) should also be considered. Key investigations should aim to confirm the diagnosis and identify the underlying cause.<br>Blood Routine Test: Essential to assess for leukocytosis (indicative of bacterial infection) or lymphocytosis (suggesting viral etiology). Inflammatory markers (e.g., ESR, CRP) may support systemic inflammation.<br>Ocular Ultrasound: A non-invasive, first-line imaging modality to evaluate the lacrimal gland for enlargement, echotexture changes, and rule out abscess formation. It is particularly effective for superficial structures like the palpebral lacrimal gland, which is already clinically evident.<br>Orbital CT/MRI: While CT provides detailed orbital imaging and is superior for detecting deeper orbital involvement or abscesses, it involves radiation exposure. MRI offers superior soft-tissue resolution but is less accessible. Given the localized findings and absence of deeper orbital signs (e.g., proptosis, optic nerve compromise), ultrasound suffices for initial evaluation.<br>Thyroid Function Tests: Less relevant here, as there are no signs of thyroid eye disease (e.g., proptosis, lid retraction, EOM restriction).<br>Conclusion: The combination of blood routine test (to identify infection/inflammation) and ocular ultrasound (to confirm lacrimal gland pathology) provides the most appropriate and efficient initial investigations for definitive diagnosis. Orbital CT/MRI may be reserved for atypical or refractory cases.<br>Answer: A. Blood routine test + ocular ultrasound |
| **Response (Gemini 2.0 PRO)** | Let's analyze this case carefully to determine the best investigations for a definitive diagnosis.<br><br>**Key Findings and Differential Diagnosis:** |

* **Bilateral eyelid swelling, orbital pain, S-shaped curvature of upper eyelids, erythematous and swollen skin, tenderness localized to the lateral upper eyelids, swollen palpebral lacrimal glands:** These strongly suggest inflammation of the lacrimal glands (dacryoadenitis). The bilateral presentation is crucial. The S-shaped curve is characteristic of lacrimal gland enlargement.
* **Lateral bulbar conjunctival congestion and edema (+):** Consistent with adjacent inflammation.
* **Normal visual acuity, transparent corneas, unremarkable fundus:** Rules out significant intraocular involvement like uveitis, optic neuritis, or other conditions directly affecting the retina or optic nerve. This makes acute, severe infections impacting vision less likely.
* **Recent cold:** A viral upper respiratory tract infection (URTI) can be a trigger for certain types of dacryoadenitis.
* **No lymphadenopathy:** This makes a widespread, systemic infectious process less likely, but doesn't rule it out completely.
*Normal Temperature: Not Febrile.

Given this clinical picture, the primary differential diagnoses include:

1. **Idiopathic Orbital Inflammation (IOI), formerly known as Orbital Pseudotumor, specifically dacryoadenitis:** This is a diagnosis of exclusion. It's a non-specific inflammatory condition that can affect any part of the orbit, including the lacrimal gland.
2. **Infectious Dacryoadenitis:** This could be viral (especially given the history of a recent cold) or, less likely given the lack of fever and bilateral presentation, bacterial.
3. **Systemic Inflammatory Diseases:** Conditions like sarcoidosis, granulomatosis with polyangiitis (GPA, formerly Wegener's), IgG4-related disease, and rarely, thyroid eye disease (though less typical without other thyroid eye signs) can cause dacryoadenitis.
4. Other very rare possibilities (lymphoma, etc).

**Why each option is correct/incorrect:**

* **A. Blood routine test + ocular ultrasound:** A blood routine test (specifically a complete blood count, CBC, and erythrocyte sedimentation rate, ESR, or C-reactive protein, CRP) is essential to assess for general inflammation and rule out obvious infection. Ocular ultrasound can be useful to visualize the lacrimal gland and assess for abscess formation, but it's not the *most* important next step for *definitive* diagnosis. It would be supportive, but not definitive.
* **B. Orbital CT:** A CT scan is a good imaging modality, but *MRI is superior for soft tissue detail.* CT is better for bony detail. While CT can show lacrimal gland enlargement, it may not differentiate the underlying cause as well as MRI.
* **C. MRI:** **This is the BEST answer.** An MRI of the orbits *with contrast* is the imaging modality of choice for evaluating dacryoadenitis. It provides excellent soft tissue detail, allowing visualization of the lacrimal gland and surrounding structures. It can help differentiate between inflammatory conditions, infection (though less likely based on presentation), and other masses. Crucially, MRI can better characterize the inflammatory process and help distinguish

| | between IOI, sarcoidosis, and other etiologies than CT. Contrast enhancement patterns are particularly helpful.<br>*   **D. Thyroid function tests:** While thyroid eye disease *can* rarely present with isolated lacrimal gland involvement, it's much less likely given the other, more prominent features of dacryoadenitis. Thyroid function tests (TSH, free T4) should be considered, but they are *not* the *most crucial* initial investigation for reaching a definitive diagnosis *in this specific presentation*. It's part of a broader workup, but not the priority.<br>*   **E. Blood routine test:** As mentioned in (A), a blood routine test (CBC, ESR/CRP) is *essential*, but not *sufficient* for a definitive diagnosis. It helps assess the presence and degree of inflammation and potential infection, but it doesn't pinpoint the *cause* of the dacryoadenitis.<br><br>**Conclusion and Definitive Diagnosis Strategy:**<br><br>The most important next step for a *definitive* diagnosis is an **MRI of the orbits with contrast (C)**. This is followed closely in importance by a **blood routine test (A, E - but A is better because it includes ultrasound)**. Thyroid function tests (D) are part of the broader workup, but lower priority than MRI and blood work. CT (B) is a reasonable imaging modality but is less sensitive for soft tissue characterization than MRI.<br><br>The definitive diagnostic approach would involve:<br><br>1.  **MRI of the orbits with contrast:** To characterize the lacrimal gland inflammation and rule out other orbital pathology.<br>2.  **Blood tests:** CBC, ESR/CRP, and potentially other tests based on MRI findings and clinical suspicion (e.g., ACE for sarcoidosis, ANCA for GPA, IgG4 levels).<br>3.  **Thyroid function tests:** To rule out thyroid eye disease, even though it's less likely.<br>4.  **Consideration of biopsy:** If the diagnosis remains uncertain after imaging and blood work, a biopsy of the lacrimal gland may be necessary, particularly to rule out malignancy or specific inflammatory conditions.<br><br>Therefore, the best single answer from the provided options is **C. MRI**. |
| **Response (OPENAI o1)** | Based on the clinical presentation and examination findings, the patient is exhibiting signs of lacrimal gland enlargement or pathology. The lacrimal gland is located in the superolateral part of the orbit, and its involvement can present with swelling and tenderness in the lateral upper eyelid, as seen in this patient. Bilateral swelling and recent history of a cold suggest a possible inflammatory process, such as acute dacryoadenitis, which can be viral in origin.<br><br>To achieve a definitive diagnosis, it is crucial to obtain detailed imaging of the orbital and lacrimal gland structures. While both CT and MRI scans can provide imaging of the orbit, MRI is superior in differentiating soft tissue structures and masses within the orbit due to its high contrast resolution. MRI can help distinguish between inflammatory conditions, neoplastic processes, and other pathologies |
| | affecting the lacrimal gland. It can provide detailed images of the lacrimal gland's size, shape, and any masses present, which is essential for accurate diagnosis and appropriate management.

Therefore, the most appropriate investigation for a definitive diagnosis in this case is an MRI.

**Answer:** C. MRI |
| **Response (OPENAI o3-mini)** | The patient's presentation, including bilateral eyelid swelling, orbital pain, erythematous and swollen skin, and tenderness, suggests a possible inflammatory or infectious condition affecting the eyelids and lacrimal glands. Additionally, the presence of swollen palpebral lacrimal glands and lateral bulbar conjunctival congestion could indicate a condition such as dacryoadenitis (inflammation of the lacrimal gland), which can be caused by infections or systemic conditions.

Given the absence of superficial lymph node enlargement, normal visual acuity, and a history of a recent cold, the primary concern seems to be an inflammatory process, potentially related to viral infection (e.g., mumps or an upper respiratory viral infection) or an autoimmune condition. Thyroid disorders, particularly thyroid eye disease, could also present with similar symptoms, especially with orbital involvement.

**Correct answer: D. Thyroid function tests**

Thyroid disease, particularly thyroid eye disease (Graves' orbitopathy), can cause orbital inflammation and eyelid changes. In this case, thyroid function tests are warranted to assess for any underlying thyroid dysfunction that might explain the symptoms. If thyroid disease is ruled out, further imaging may be needed for a definitive diagnosis. |